# ROBOT LOCALIZATION AIDED BY QUANTUM ALGORITHMS

Unai Antero[1,2], Basilio Sierra[2], Jon Oñativia[1], Alejandra Ruiz[1], and Eneko Osaba[1]

[1]TECNALIA, Basque Research and Technology Alliance (BRTA), Mikeletegi Pasealekua, 7. Donostia, 20009. Spain
[2]Robotics and Autonomous Systems Group (RSAIT), University of the Basque Country (UPV-EHU). Manuel Lardizabal 1, San Sebastian, 20018. Spain

May 1, 2025

## ABSTRACT

Localization is a critical aspect of mobile robotics, enabling robots to navigate their environment efficiently and avoid obstacles.

Current probabilistic localization methods, such as the Adaptive-Monte Carlo localization (AMCL) algorithm, are computationally intensive and may struggle with large maps or high-resolution sensor data. This paper explores the application of quantum computing in robotics, focusing on the use of Grover's search algorithm to improve the efficiency of localization in mobile robots. We propose a novel approach to utilize Grover's algorithm in a 2D map, enabling faster and more efficient localization.

Despite the limitations of current physical quantum computers, our experimental results demonstrate a significant speedup over classical methods, highlighting the potential of quantum computing to improve robotic localization. This work bridges the gap between quantum computing and robotics, providing a practical solution for robotic localization and paving the way for future research in quantum robotics.

## 1 Introduction

Localization is a vital aspect of mobile robotics, enabling robots to navigate their environment efficiently and avoid obstacles. Without localization, mobile robots would not be able to determine their position and orientation, making it difficult to plan a path or make informed decisions about their movement ([29]). Localization allows mobile robots to create an internal map of their environment, which is essential for tasks such as surveying, manipulation, inspection, and delivery ([21]). In fact, based on the knowledge of a previously built map, localization is what enables mobile robots to perform tasks autonomously, allowing them to make the right actions and movements without human intervention.

The quality of localization is heavily dependent on the generation of accurate maps, which is a computationally intensive task. Probabilistic localization methods, such as the Adaptive Monte Carlo localization (AMCL) algorithm, have been widely used in mobile robotics due to their accuracy and robustness ([25]). However, these methods can be computationally demanding, especially when dealing with large maps or high-resolution sensor data. AMCL, in particular, uses a combination of sensor data and prior map knowledge to determine the probable location of a robot on a given map, but its computation complexity is proportional to the area of the grid of the map ([6]).

Recently, the integration of light detection and ranging (LiDAR) sensors has improved the accuracy of localization methods, but computational requirements remain a challenge ([21]). To overcome this challenge, researchers have started exploring the application of quantum computing in robotics, which has the potential to revolutionize the field by providing faster and more efficient computation ([27]).

Quantum computing is a new paradigm for computing that has shown significant promise in various fields, including cryptography, optimization, and machine learning. Its application in robotics is still in its early stages, but it has the potential to provide breakthroughs in areas such as localization, planning, and control ([30]). Quantum-inspired algorithms, such as Grover's search, have been shown to provide computational speedups in certain tasks, making them an attractive solution for robotic applications ([35]). Different proposals have been subject to research, such as planning ([13]), but new approaches that benefit from the quantum advantage have not yet been fully explored.

This paper aims to evaluate the introduction of quantum computing in robotics, focusing on the application of quantum algorithms in robot localization. Specifically, it explores the use of Grover's search algorithm in AMCL to improve the efficiency of localization in mobile robots. Although previous studies have explored the application of quantum planners in robotic navigation, their main focus has been on the use of quantum algorithms to improve decision-making optimization rather than integrating quantum computing into the localization module itself ([14]). Our approach aims to bridge this gap by introducing a pragmatic approach to robot localization that can be effectively implemented in real-world scenarios.

### 1.1 Research Objectives

Quantum computing has the potential to bring significant benefits to robotics science by enhancing computational power, optimizing algorithms, and enabling new types of processing that are not possible with classical computers ([28]). Quantum algorithms leverage the principles of quantum mechanics to perform certain computations more efficiently than classical algorithms.

Grover's algorithm is one of the most famous quantum algorithms introduced by Lov Grover in 1996 ([16]). It is a quantum search algorithm that finds an element in an unsorted database of $N$ items in $O(\sqrt{N})$ time, compared to classical algorithms that require $O(N)$ time. Grover's algorithm is suitable for applications involving unsorted database search, cryptography, and combinatorial optimization, where it can provide a quadratic speedup over classical algorithms ([18]).

In this paper, we explore the application of quantum computing in robotics, focusing on the improvement of localization algorithms using quantum search techniques. We propose a novel approach to utilize Grover's algorithm in a 2D map, enabling faster and more efficient localization. Our goal is to evaluate the potential benefits of quantum computing in robotics and to provide a practical solution for robotic localization.

Our work demonstrates the potential of quantum computing to improve robotic localization and highlights the practical applications of quantum algorithms in robotics. We show that our proposed approach can provide a significant speedup over classical methods and discuss the implications of our results for future research in quantum robotics.

The remainder of the paper is organized as follows. We begin by describing the map encoding process, including the calculation of costmaps and the translation of localization into a search problem. We then discuss the limitations of classical search algorithms and how quantum search improves upon these methods. Next, we introduce Grover's algorithm and its application in the localization problem.

We then present our proposed approach for using Grover's algorithm in a 2D map, including the algorithm's description and its application in a costmap. We describe our experimental setup and validation procedures, including simulations and experiments on real quantum computers.

## 2 Background

Effective navigation is a fundamental capability for mobile robots, based on the successful integration of four essential components: sensor-based perception, self-localization, decision-making, and precise motion control. Among these components, self-localization is crucial, as it enables the robot to determine its position and orientation, providing a comprehensive understanding of its spatial disposition within its environment.

Robot localization entails the estimation of a robot's pose, comprising its position (defined by three-dimensional Cartesian coordinates) and orientation (characterized by three-dimensional Euler angles or quaternions). This process is vital for mobile robots to interact efficiently with their environment and achieve their goals.

Although classical-computing localization methods have shown evident and significant progress, the increasing complexity of robotic applications demands more efficient and robust techniques. The emergence of quantum computing offers possible new opportunities to enhance localization methods, with the promise of enabling faster processing and/or more accurate positioning. In the near future, quantum computing is expected to revolutionize the field of robotics in general (and localization in particular), enabling robots to navigate and interact with their environment.

The next subsections explore the current state of the art in two aspects which are required for the approach presented in this paper: on the one hand, how map encoding is calculated, and on the other hand, how the quantum advantage can be used to speed up the search algorithms.

## 2.1 Map encoding

Map encoding (encoding the map information into a form that can be manipulated by the robot's CPU) is a required step in any localization technique. In recent decades, robotic research has explored and developed different mapping technologies ([37]), ([26]), being the "occupancy grid" approach one of the dominant paradigms in the robotics field ([15]).

The core concept of the occupancy grid is to model the environment as a regularly spaced grid, where cells contain a certain numeric value that indicates the probability that the cell is traversable (measuring in certain way whether the cell is occupied by an obstacle, the cell has a nearby obstacle, or is fully traversable) [36].

In some naive approaches, the cost map utilized for computing the occupancy grid is binary (a cell cost is not zero when the cell is occupied by an obstacle and zero otherwise) [12], but usually in cost map planning, the cost of each cell ranges from 0 to $\inf$, where 0 corresponds to a fully traversable cell, $\inf$ corresponds to an obstacle (untraversable), and any finite value in between corresponds to some degree of traversability (or some other cost function) [20]. In this paper, we follow the second approach, proposing a simple algorithm (simple enough so it can be used in a quantum computer) that creates a costmap that tries not only to identify the specific location of obstacles, but also the presence of nearby obstacles.

The proposed algorithm for the cost map has three steps: a first step where for every obstacle its influence in nearby cell is computed is presented, a second step where for each cell a sum of such influences is calculated (considering the distance to the obstacles), and a final third step where the cost values are normalized to compact the information (use less bits in its representation). It should be noted that distances are measured using the Manhattan distance [34].

These steps are presented in Algorithm 1:

---

**Algorithm 1** Simple costmap algorithm, assuming InitialCostValue=4

**Require:** $B_1 \ldots B_K$ as a list of K obstacles
**Require:** $C_1 \ldots C_N$ a a list of N cells in a $n \times n$ grid, being $N = n \times n$
**Require:** $Q$ as the number of bits to be used in the normalization

    // Initialize initial cost for each cell
    *Costmap value of cell* $C_1 \ldots C_N \leftarrow 0$
    **for** $k \leftarrow 1$ to $K$ **do**
        *InitialCostValue* $\leftarrow n$
        **for** $z \leftarrow 1$ to $N$ **do**
            *DistanceToObstacle* $\leftarrow$ *distance to* $B_k$ *in steps*
            *PreviousCostmapValue* $\leftarrow$ *Costmap value of cell* $C_z$
            *Increment* $\leftarrow$ *InitialCostValue*$/(2^{StepsToObstacle-1})$
            *Costmap value of cell* $C_z \leftarrow$ *PreviousCostmapValue + Increment*
        **end for**
    **end for**

    // Now normalize values in cells to use $Q$ bits
    *MaxValue* $\leftarrow$ *max. value in cells* $C_1 \ldots C_N$
    *MinValue* $\leftarrow$ *max. value in cells* $C_1 \ldots C_N$
    **for** $z \leftarrow 1$ to $N$ **do**
        *Factor* $\leftarrow$ (*MaxValue* $-$ *MinValue*)$/(2^Q - 1)$
        *Costmap value of cell* $C_z \leftarrow$ *PreviousCostmapValue/Factor*
    **end for**

---

Consider the setup shown in Figure 1.

If we consider the first step where we evaluate the costmap for the obstacles, this means:

The final costmap (the one in red) in Figure 2 shows a set of values that, in some sense, measure the impact of nearby obstacles in each cell. Normalization has been used to optimize the encoding of values to avoid wasting bits.

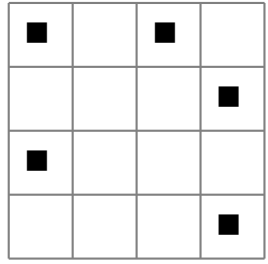

Figure 1: Sample 4x4 grid with obstacles (in black)

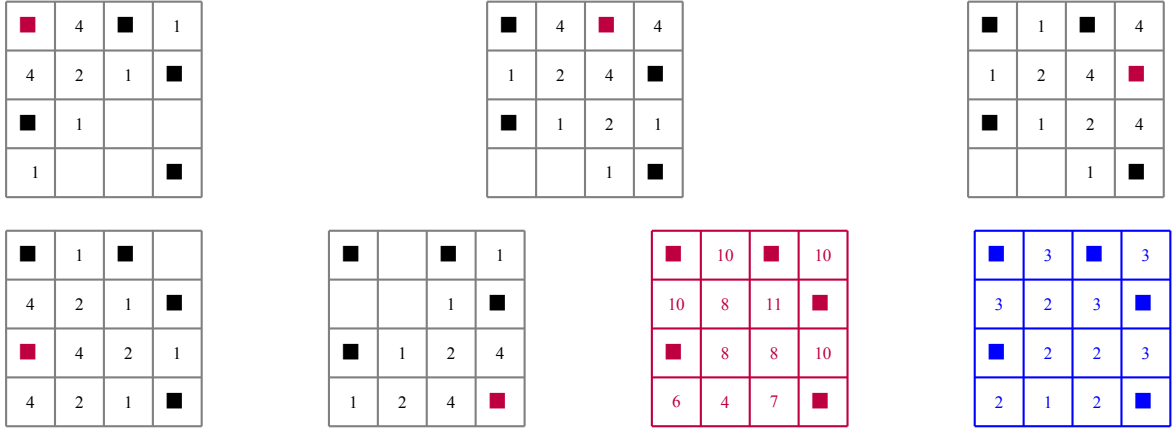

Figure 2: Costmap incremental calculation steps for each of the five considered obstacles (shown in red in each step), final value of costmap (in red) and normalized values (in blue)

In a real environment, sensors have limitations, and their reach is limited. For our analysis here, if we consider that the robot sensors can perceive their environment up to a distance of two cells (sensors cannot perceive objects beyond two steps in the grid), using the Algorithm 1, we can represent what the robot would perceive in some position on the grid.

We show in Figure 3 what the robot would perceive (the robot is marked in the figure with a ® mark), and the normalized costmap is calculated:

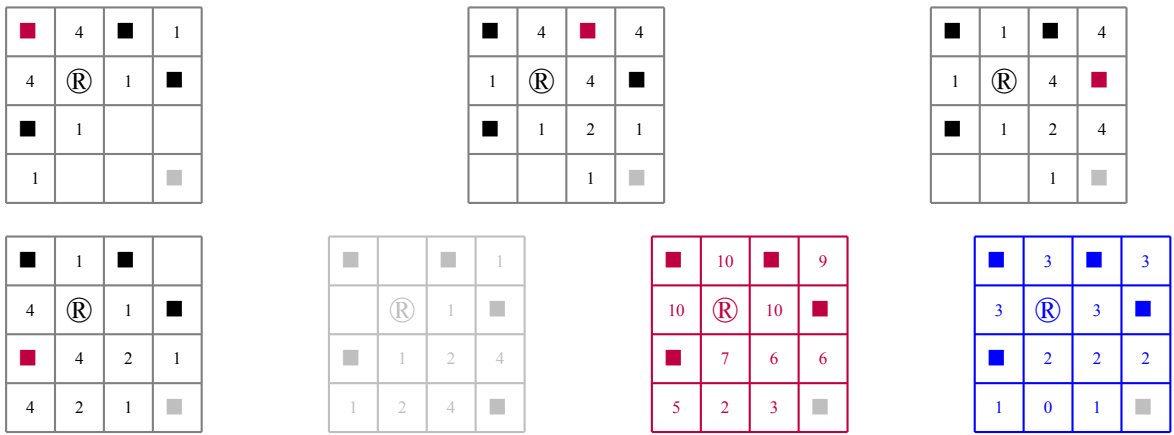

Figure 3: The robot is into an unknown location. Figure shows costmap calculation steps for each detected obstacle (marked in red). In gray, obstacles which are beyond detection, hence they are not used in the calculation. Final value of costmap (in red) as the sum of all values for detected obstacles, and normalized values (in blue)

Now, the localization problem transforms into trying to match what the robot is perceiving. This match is performed using the previously stored costmap of the environment.

In the context of robot localization, a match between a robot's perception and a previously stored costmap can be defined as a measure of similarity or correlation between the two. For the sake of simplicity, in the scope of this paper a match between a robot's perception and a previously stored costmap is established when the perceived encoded value is identical to the stored encoded value (no tolerance or margin of error is allowed, meaning the values must be exactly equal for a match to be declared).

Figure 4 shows graphically what the robot is perceiving (left) and what was previously encoded and stored (right).

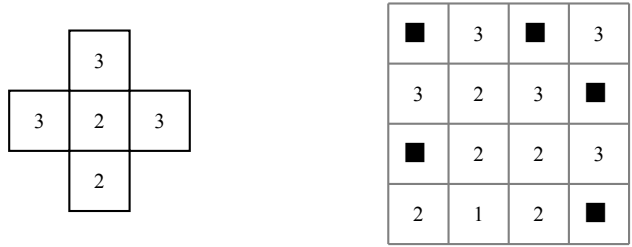

Figure 4: The robot localization requires to match the perception costmap (left), with the previously encoded and stored costmap (right)

In other words, if we split the problem into horizontal and vertical matching (rows and columns), we can convert the problem into a problem of finding a certain substring in a longer string: "323" in "33333233332232123" (rows) and "322" in "3332322133223333" (columns) (in the normalization we consider the value of the obstacles as the maximum value in the normalization).

String searching algorithms, also known as string matching algorithms, are a crucial type of string algorithms designed to locate the position of one or more specific strings (referred to as patterns) within a larger string or block of text, which are commonly used in many different areas ([39]): from daily activities (such as searching for a text in an email), to language translation, data compression, or bioinformatics (when trying to find a certain DNA sequence ([33])), etc.

In the case presented in the previous paragraphs (finding "323" in a longer text), a simple and straightforward (but naive) algorithm would be to create a loop and compare each element in the occupancy grid array with the target data. Starting with the first character of the larger string (the haystack), we compare it with the smaller string (the needle). If there is no match, we move to the second character and repeat the process, continuing this way for each position. Typically, only a few characters must be checked at each position before determining whether it is incorrect, so on average, this approach requires $O(n + m)$ steps, where n is the length of the haystack and m is the length of the needle. However, in the worst-case scenario, such as when searching for a pattern like "aaaab" within "aaaaaaaab", the process can take up to $O(nm)$ steps.

Different algorithms have tried to optimize the time required for substring search in classical computing: from the optimized algorithms created in the seventies (such as the Knuth-Morris-Pratt Algorithm ([24]), or the Boyer-Moore Algorithm ([11])), to the new approaches in the eighties (with Yaeza-Bates ([9]) and many others) and later. But this problem still requires $O(n+ m)$ time if tackled classically (using classical computation).

## 2.2   Using the quantum advantage in the localization problem

As a complementary way of dealing with substring search, as any algorithmic improvements in this problem will result in great impacts in many areas, since 2000 quantum algorithms have been designed for the string matching problem ([31]).

The starting point was the quantum Grover algorithm that searches for an element in a unordered database in $O(\sqrt{n})$ time. As indicated by [19], finding whether the pattern matches somewhere in the text is equivalent to searching in an unordered database; so we can have a quadratic speed-up compared to classical methods.

With the increasing demand for robust and accurate robot localization, researchers have explored innovative approaches to address the limitations of traditional methods. In recent years, the intersection of robotics and quantum computing has opened new avenues for solving complex problems.

Inspired by the principles of quantum mechanics, we propose a novel approach to robot localization that takes advantage of the power of quantum algorithms. Specifically, we draw inspiration from the Grover algorithm, a quantum search algorithm known for its exponential speedup over classical search algorithms.

We begin by discretizing the environment into a rectangular grid consisting of $n \times n$ cells, each representing a small region of space. This grid serves as a framework for representing the robot's surroundings and can be used to create an occupancy grid. In this occupancy grid, each cell is assigned a measurement derived from the robot's onboard sensors (such as lidar, sonar, infrared readings, etc.). The measurement represents the likelihood that the cell is occupied by an obstacle or is free space. This process is repeated as the robot explores the environment, resulting in a global annotated occupancy grid that captures the spatial layout of the environment.

When the robot is placed in an unknown position, it takes its new sensor readings and compares them with the annotated occupancy grid. To accelerate the search for the robot's location, instead of using a classical method (such as using a particle filter), we employ a quantum search algorithm inspired by Grover's algorithm. This algorithm efficiently searches the occupancy grid to find the cell that best matches the current sensor readings, effectively localizing the robot within the environment. Taking advantage of quantum computing, the search process is significantly accelerated, allowing for fast and accurate robot localization even in complex and dynamic environments.

As an example, let's use a simple environment divided into a 4x4 grid as shown in the Figure 5:

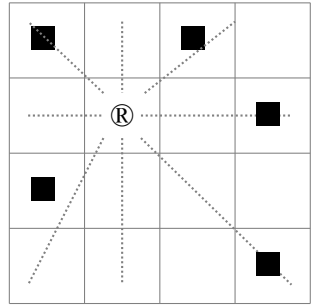

Figure 5: The black boxes mark the obstacles, and ® marks the robot position

In this updated figure, we show the robot, the obstacles and lidar beams (marked as dotted lines).

In order to be able to localize the robot in the map, two things are needed: some way to encode the map and some way to use the information measured from the lidar to locate the robot in the encoded map.

### 2.2.1 Grover's quantum searching algorithm

Grover's quantum searching algorithm is usually described ([16]) in terms of the iteration of a compound operator Q of the form ([23])

$$Q = -HI_0 HI_{x_0} \quad (1)$$

On a starting state $|\psi_0\rangle = H|0\rangle$. Here $H$ is the Walsh-Hadamard transform and $I_0$, $I_{x_0}$ are suitable inversion operators. The operator $-HI_0 H$ was originally called a "diffusion" operator ([16]) and later interpreted as "inversion in the average" ([17]).

The search problem is often phrased in terms of an exponentially large unstructured database with $N = 2^n$ records, of which one is specially marked. The problem is to locate the special record. Elementary probability theory shows that classically if we examine $k$ records then we have probability $k/N$ of finding the special one so we need $O(N)$ such trials to find it with any constant (independent of $N$) level of probability. Grover's quantum algorithm achieves this result with only $O(\sqrt{N})$ steps (or more precisely $O(\sqrt{N})$ iterations of $Q$ but $O(\sqrt{N} \log N)$ steps, the $\log N$ term coming from the implementation of $H$.)

The search problem may be more accurately phrased in terms of an oracle problem, which we adopt here.

We will replace the database by an oracle which computes an $n$ bit function $f : B^n \to B$ (where $B = \{0, 1\}$). It is promised that $f(x) = 0$ for all $n$ bit strings except exactly one string, denoted $x_0$ (the "marked" position) for which $f(x_0) = 1$. Our problem is to determine $x_0$. We assume as usual that $f$ is given as a unitary transformation $U_f$ on $n + 1$ qubits defined by

$$U_f |x\rangle |y\rangle = |x\rangle |y \oplus f(x)\rangle \quad (2)$$

Here, the input register $|x\rangle$ consists of $n$ qubits as $x$ ranges over all $n$ bit strings and the output register $|y\rangle$ consists of a single qubit with $y = 0$ or 1. The symbol $\oplus$ denoted the addition modulo 2 ("modulo" is the remainder of a division operation between two numbers).

The assumption that the database was unstructured is formalized here as the standard oracle idealization that we have no access to the internal workings of $U_f$ – it operates as a "black box" on the input and output registers.

In this formulation, there is no problem with the access to $f(x)$ for any of the exponentially many values of $x$, and indeed we may also readily query the oracle with a superposition of input values.

Instead of using $U_f$, an equivalent operation is used. This operation is denoted $I_{x_0}$ on $n$ qubits. It is defined by

$$I_{x_0} |x\rangle = \begin{cases} |x\rangle & \text{if } x \neq x_0 \\ -|x_0\rangle & \text{if } x = x_0 \end{cases} \qquad (3)$$

i.e. $I_{x_0}$ simply inverts the amplitude of the $|x_0\rangle$ component. If $x_0$ is the $n$ bit string $00\ldots0$ then $I_{x_0}$ will be written simply as $I_0$.

Our searching problem becomes the following: we are given a black box that computes $I_{x_0}$ for some $n$ bit string $x_0$ and we want to determine the value of $x_0$.

Given the black box $I_{x_0}$, Grover's algorithm 2 states that in order to locate x0 (in an unordered list), we should follow the algorithm:

---
**Algorithm 2** Grover Algorithm
---
    Initialize the state $|s\rangle$
    (to a uniform superposition of all possible states)
    **for** $z \leftarrow 1$ to $O(\sqrt{N})$ **times do**
        $|s\rangle \leftarrow$ Apply Grover's oracle $U_f$ to the state $|s\rangle$
        $|s\rangle \leftarrow$ Apply Grover's diffusion operator $U_s$ to the state $|s\rangle$
    **end for**

    Measure the state $|s\rangle$ to gather the solution(s).
---

Figure 6 shows the main steps of the Grover algorithm, with an initial preparation step, one block that contains the Oracle and the diffuser (which must be repeated), and a final measurement step.

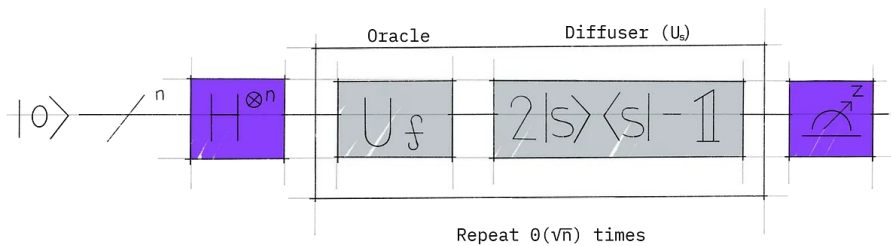

Figure 6: Steps in Grover algorithm - Illustration from https://learn.qiskit.org/course/ch-algorithms/grovers-algorithm

With this algorithm at hand, we can try to check if it fits our localization problem. If we take Figure 4 and write the numbers in binary form, we obtain Figure 7

We can try to check if this algorithm could be used to find the position of a certain substring (obtained by the sensors of the robot) in a longer string (where the environment map is encoded). For example, we can use Qiskit to implement Grover's algorithm for such detection. They key here is to be able to define an Oracle based on a function that achieves $f(x_0)=1$, and then apply the Algorithm 2.

If we use "10110001" as a certain row where we look for the pattern "01", and use Grover's algorithm, we obtain the results shown in Figure 8:

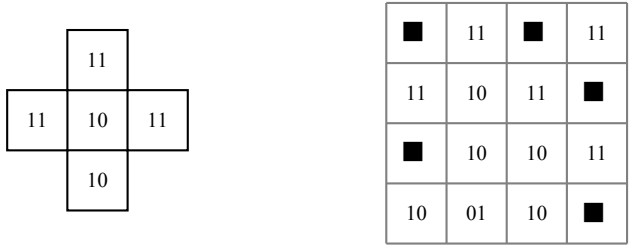

Figure 7: The robot localization should be done trying to match the perception costmap (left), with the previously generated costmap (right)

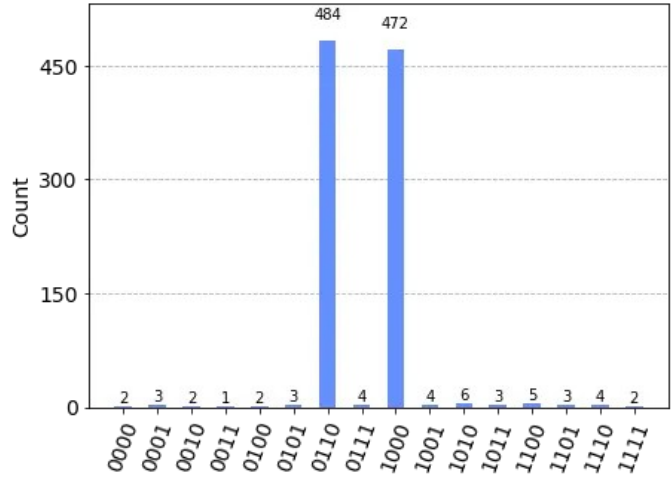

Figure 8: substring location using Grover's algorithm, programmed in Qiskit and simulated

## 3 Proposed approach

Grover's algorithm offers a powerful quantum computing approach to searching problems, and it can be adapted for robot localization within a known map dramatically reducing the time complexity associated with finding the correct position. In a typical robot localization problem, the goal is to identify the robot's position within a predefined environment. This environment can be represented as a map with a finite number of discrete locations, each representing a possible position where the robot might be.

This paper explores the application of quantum technologies to robot localization problems, specifically by extending Grover's algorithm. Our approach goes beyond a simple implementation of Grover's algorithm, proposing a double search method: one along the horizontal axis and another along the vertical axis, enabling a comprehensive two-dimensional search within a 2D map.

The algorithm has been implemented using Qiskit ([2]), in a way that associates each candidate location on the map with a quantum state, where the correct state (actual location of the robot) is the one marked by an 'oracle' function. This oracle has been programmed to recognize and mark the actual location by matching the candidate state with the real-time data collected from the robot sensors. The oracle then enhances the amplitude of the state that represent the correct location. By repeatedly applying the Grover operator, the probability amplitude of the correct position is amplified with each iteration, making it more likely to be measured upon observation.

In practical terms (if real quantum computing achieves enough capacity), this quantum-enhanced localization approach could be particularly beneficial in environments with high location ambiguity or large search spaces, such as a warehouse with many aisles or complex navigation zones. In principle, leveraging quantum computing's ability to quickly process great amounts of data (exponentially faster than classical computers), robots and autonomous vehicles can quickly determine their precise location and orientation, even in situations where GPS signals are weak or unavailable.

Although the use of Grover's algorithm in real-time robotic applications would require specialized quantum hardware and the integration of quantum-classical computing elements, the potential for accelerated search could be transformative, especially in scenarios where rapid localization is critical for decision-making and navigation.

That is why in this paper we propose using Grover's quantum search oracle to perform a double localization on a 2D map: one for the horizontal axis and another for the vertical axis, as shown in Figure 9

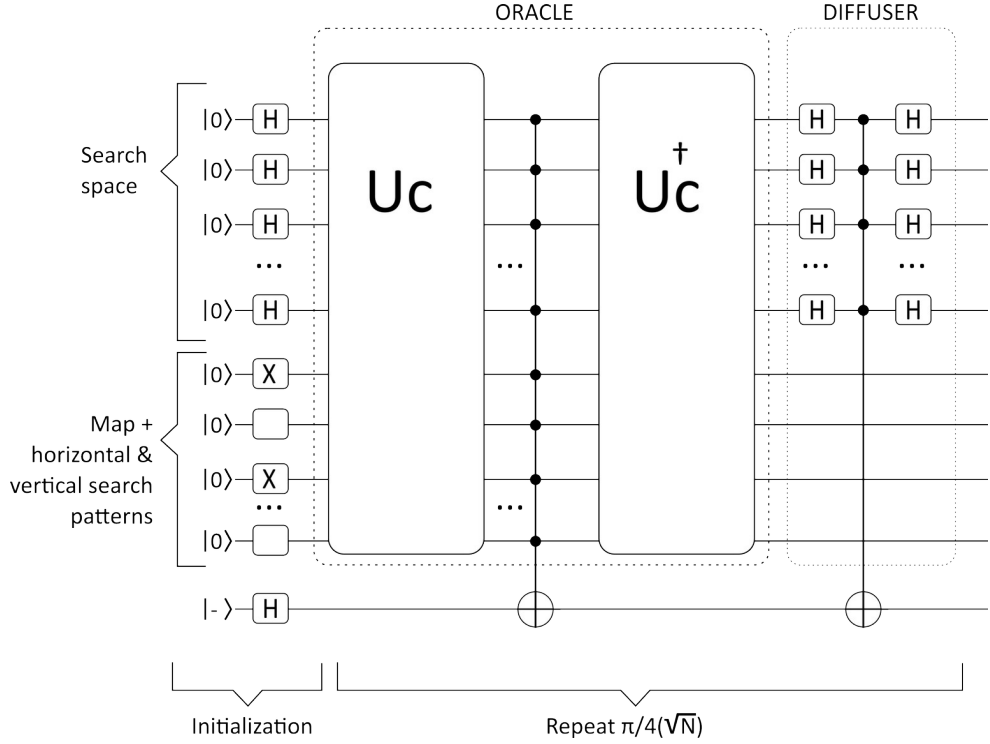

Figure 9: Block diagram of the proposed approach

Conceptually, the approach is quite simple: encode the 2D map in a quantum register, encode the robot's perception in the vertical and horizontal axes in two quantum registers, and follow Grover's algorithm using two consecutive oracles (one for the vertical and another one for the horizontal axes).

We left aside other current approaches for pattern matching, like the use of Parametric Probabilistic Quantum Memory ([32]) (PQM, a data structure that computes the Hamming distance from a binary input to all binary patterns stored in superposition on the memory), as they require the substring to be searched to be of the same length as the strings in the stored data.

Additionally, it must be noted that in the scope of this paper, a complete software architecture based on Qiskit has ben developed to guarantee that the generated code can be effectively used (up to certain scale) as an alternative to current robot localization methods.

### 3.1 Algorithm

The proposed algorithm is presented below as Algorithm 3:

**Algorithm 3** Grover search for a 2D search

· Initialize the state $|s\rangle$ (to a uniform superposition of all possible states)
· Encode the map (each of its grid) into a $|m\rangle$
· Encode the robot's perception in the x axis (rows) in a $|r\rangle$
· Encode the robot's perception in the y axis (columns) in a $|c\rangle$
· Prepare the Oracle $U_f$ that for every specific (r,c) in $|s\rangle$, returns 1 if the map's grid's value is equal to what the robot perceived (and is stored in $|r\rangle$ and $|c\rangle$)

**for** $z \leftarrow 1$ to $O(\sqrt{N})$ **times do**
    $|s\rangle \leftarrow$ Apply Grover's oracle $U_f$ to the state $|s\rangle$
    $|s\rangle \leftarrow$ Apply Grover's diffusion operator $U_s$ to the state $|s\rangle$
**end for**

Measure the state $|s\rangle$ to gather the solution(s), and extract the horizontal and vertical coordinates from the solution.

The following pictures describe how the algorithm 3 works:

Let's take the following situation described in Figure 10:

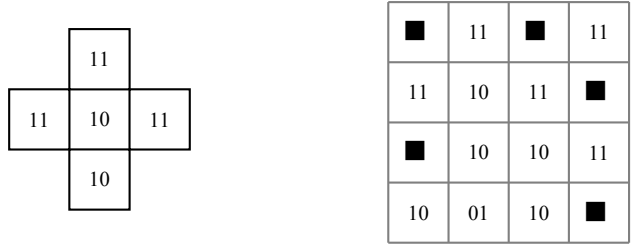

Figure 10: The perception costmap (left) has 3 horizontal elements, and 3 vertical elements. The map (the previously generated costmap, in the right) is a 4x4 grid of cells (each cell has two qubits)

The search space $|s\rangle$ is initialized in a superposition state using Hadamard gates. The horizontal perception costmap has to be encoded in a register $|r\rangle$ with 3 qubits. The vertical perception costmap has to be encoded in a $|c\rangle$ with 3 qubits. Finally, the map will be encoded in $|m\rangle$, a register with 32 qubits (4x4 cells, each one holding two qubits).

Figure 11: Considering (0,0) as the upper left cell, for coordinates (1,1) if the map's cell values (in black) match the values for the horizontal (blue) and vertical (red), the Oracle's output should be '1', and '0' otherwise. In other words: the algorithm should look for [11,10,11] horizontally, and [11,10,10] in vertical.

The code generated in this paper ([7]) can be configured for any map size, any number of qubits per cell, and any row or column pattern. In this case (with a 4x4 map with 2 qubits per cell and horizontal and vertical patterns with size 3 (using 2 qubits per cell), the Oracle is automatically generated. Such an oracle (in this case) has a possible search space of: [(1,1), (1,2), (2,1), (2,2)], and would return '1' in the output qubit for position: (1,1).

# 4 Experimental Setup and Validation

In the preceding sections, we introduced a novel application of quantum computing to the problem of robot localization, proposing a quantum algorithm designed to efficiently solve the localization problem. To validate the feasibility and effectiveness of this quantum approach, this section presents a comprehensive experimental setup and validation framework. We outline the methodology employed to evaluate the performance of our quantum algorithm in estimating a robot's position and orientation within a given environment and present the results obtained from a series of simulations and experiments. These experiments aim to assess the accuracy, robustness, and scalability of our quantum algorithm, and demonstrate its potential advantages over classical methods for robot localization. The remainder of this section details the experimental design, performance metrics, and results, offering a thorough analysis of the proposed quantum approach to robot localization and its potential applications in robotics and autonomous systems.

To validate the proposed quantum algorithm for robot localization, we have developed a software framework consisting of two complementary modules: an auxiliary module for configuration and a main module.

The first module is an auxiliary configuration tool developed using Pygame ([1]) (a Python library designed for creating video games and multimedia applications). This tool enables users to define and customize the parameters of the localization problem, including the 2D map, robot properties, and algorithmic settings. The output of the tool is a standardized input file that serves as the basis for the second module.

This configuration utility can be used to:

- Create a map with a size $n \times n$, being $n \geq 2$
- Define location of obstacles in the map
- Define location of the robot in the map
- Define number of qubits to be used in each cell of the grid (allowing to configure that number with any integer $\geq 1$)

Once launched, the application shows a simple user interface where, using arrow keys, it is possible to move the robot to any cell in the map.

This user interface is shown in Figure 12:

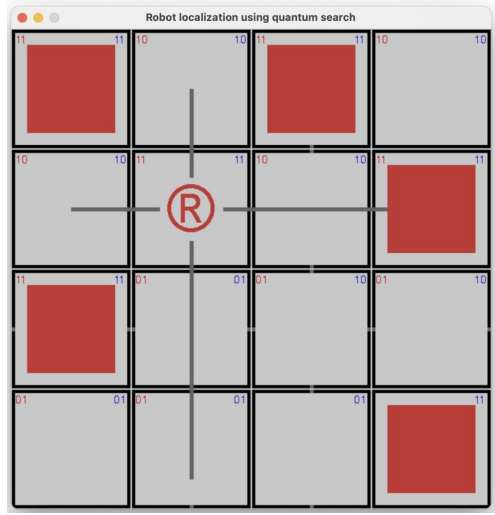

Figure 12: Configuration tool developed using Pygame. Red squares mark obstacles, ® marks the robot position. Vertical and horizontal lines represent the robot's sensor. In each cell, number in red marks the sensor costmap, and the number in blue the global costmap

Using this user interface, different setups (described in the following subsections) have been prepared and tested. These setups cover different situations: from small grids that were simulated locally in a PC, to more complex grids that had to be simulated in real quantum computers.

On the other hand, within the scope of this paper, we have also implemented the algorithm described in 3. This main module, built using the Qiskit quantum development environment, implements the quantum algorithm for robot localization in a 2D map.

Designed with modularity and flexibility in mind, this module takes the configuration file generated by the auxiliary tool as input, allowing for seamless integration and adaptability to various problem settings.

By decoupling the problem configuration from the implementation of the quantum algorithm, our framework provides a versatile and user-friendly environment for testing and evaluating the performance of the proposed quantum algorithm for robot localization.

In any case, it should be noted that quantum computing, as a technology, is still in its infancy. It is clear that different engineering challenges in the real quantum computer disable (in a certain sense) any advantage that the proposed algorithm may present. In other words: notwithstanding the theoretical validity of the proposed quantum approach, the practical implementation of the corresponding quantum circuit is subject to the engineering limitations of current quantum computing technology.

As a result, even if the underlying scientific principles are sound, the actual effectiveness of the approach may be constrained by the technical capabilities of existing quantum computers.

In real quantum computers, we currently face four main challenges:

- Stability of the Qubit state: The stability of a qubit's state typically lasts only a few hundred microseconds, which limits the duration of computations.
- Noisy quantum gates: The operations performed on qubits are inherently noisy, causing a loss of accuracy with every calculation.
- Measurement errors: the process of measuring a qubit's state is prone to inaccuracies.
- Limited qubit availability: The number of physical qubits available in these devices remains constrained, which poses a significant bottleneck for scalability.

Limited qubit availability impacts on the complexity of the algorithms we can use, but the other three issues directly impact the error probability of the final result. As indicated in ([22]), it is important to understand the effect of these issues and to have some estimation of the final error.

In this paper, we use data provided by the manufacturer of the real quantum computer (IBM in our case), to calculate the total error probability expected during the run of a given quantum circuit. Using the methodology (and code) proposed in ([8]), for each setup we estimate how far we are from the limits of current quantum computers (for using their code in this paper, the code has been updated to the latest version of Qiskit).

In the next subsections, we describe the experimental setups and results obtained using the software developed in this paper, which demonstrates the effectiveness of our quantum approach to robot localization (and the limits of current physical quantum computers).

## 4.1 Small grid with simple pattern match

As a first test for validation, a very simple setup has been configured: it is a 2x3 qubit grid that uses a single qubit per cell to encode the costmap.

The Pygame based interface has been used to configure such a grid, and to define the robot's position in a certain cell in the grid.

An obstacle has been placed in an arbitrary location, as seen in Figure 13. Now, the goal is to find the location in which a robot would perceive such an obstacle.

The proposed double (horizontal and vertical) Grover search with a pattern [1] in a $2 \times 3$ grid, has to deal with a search space of 6 qubits. For this setup, our algorithm requires us to use:

- The Oracle has to deal with 6 possible positions (2x3), with the expectation of a single correct solution (N=6, M=1).
- This requires 3 qubits for the search space (to encode 6 positions)
- The suggested number of repetitions is: $R = \pi/4 \sqrt{N/M} = 2$
- The total number of required qubits is 13

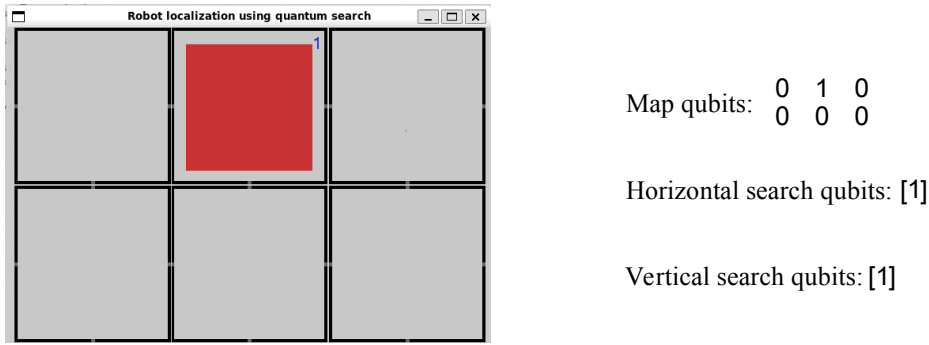

Map qubits: $\begin{matrix} 0 & 1 & 0 \\ 0 & 0 & 0 \end{matrix}$

Horizontal search qubits: [1]

Vertical search qubits: [1]

Figure 13: Simple setup with a 2x3 grid

- Size of the quantum circuit is 163 elements (with a depth of 89)

In order to validate the effectiveness of the algorithm, the Qiskit code has been tested on different systems (simulated and real):

- Using the IonQ Quantum Cloud ([5]), where it is possible to simulate circuits as if they were running on IonQ hardware, using a special noise model simulation.

- Using the Bluequbit Cloud platform ([4]), a platform specialized in the simulation of quantum circuits .

- Using a simulation based on the 'qasm' simulator, but where noise and error rates of a real quantum computer have been introduced. In this test, the error rates of IBM's "ibm brisbane" quantum processing unit weree used (a 127qubit unit, with a 2Q error (best) of 3.28e-3).

- Using a real quantum computer from IBM ([β]). Using IBM's SDK it is possible to send a quantum circuit to a real IBM quantum computer. In this test, a real "ibm brisbane" quantum processing unit was also used for the execution of the algorithm (the code was sent remotely to be executed in a job queue).

Grover's algorithm identified the correct solution: row 1 and column 2, identified by position '0' or '001' in binary (count starts from zero on the upper left corner of the grid). The probability of the correct solution was amplified in the simulated environment and in the real environment.

However, since these operations are performed in the realm of quantum mechanics, governed by the principles of superposition, entanglement, and interference, there is the possibility that the results found vary from the calculations in the ideal environment.

This could be computed due to the existence of noise, since noise refers to random errors that occur during computation.

These errors cause the qubits to lose their quantum properties and behave classically. Consequently, as a direct result, actual computations diversify from the calculated probability and from simulated equations.

Furthermore, real quantum computers operate differently from ideal simulations because of their susceptibility to noise.

In the presence of noise, the results of the computation may tend to deviate from the results in the ideal simulation, consequently causing discrepancies between the results and expectations of behavior from the quantum system.

The histograms obtained are shown in Figure 14:

It is clear that the noise in the real quantum computer masks (in a certain sense) any advantage that the proposed algorithm may present. In the tests performed, although most of the time the results are correct, sometimes the algorithm tends to fail on the real computer. In the "ibm brisbane" quantum unit that has been used, the lowest two-qubit error rate (as indicated by IBM) is 3.28e-3. Using this information, it is possible to have a rough estimate of the probability that the algorithm ran without any error (using the 'depth' as the number of layers that work one after the other):

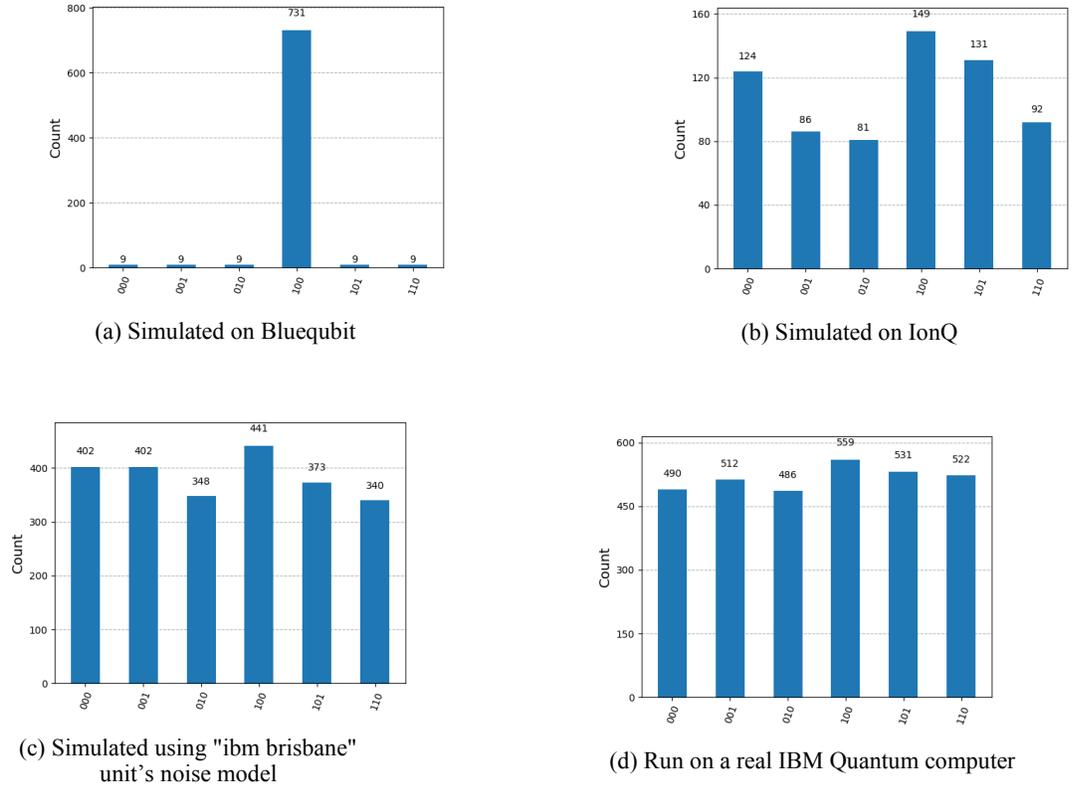

Figure 14: Histograms of the same quantum circuit obtained from simulated and real quantum computers

$$\begin{aligned}
P \text{ (at least 1 gate fails)} &= 1 - P \text{ ( all gates succeed)} \\
&= 1 - P \text{ ( one gate succeeds)}^{\text{number of gates}} \\
&= 1 - (1 - P \text{ ( gate error)})^{\text{aprox. depth of circuit}} \\
&= 1 - (1 - 3.28e - 3)^{89} \\
&\approx 25\%
\end{aligned} \quad (4)$$

This value is aligned with how the expected error is estimated according to [8] whose code has been integrated in the code of this paper and estimates that for this 2x3 grid we get:

- Probability of error due to noisy gates = 28%
- Probability due to measurement errors = 18%

### 4.2 Bigger grid, with simple pattern match

To perform a more useful validation, it is necessary to perform a simulation that is computationally more demanding, so in this setup a 5x5 grid will be used. The setup considers a 1-qubit pattern search (both horizontally and vertically), and the calculated costmap in each cell has been configured to use a single qubit. Once again, the Pygame-based interface has been used. The robot has been placed in the central cell, as seen in Figure 15.

The proposed double Grover search with a pattern [1] has to deal with a search space of 5*x*5. In order to evaluate the quantum circuit, the proposed algorithm needs:

- 5 qubits for the search space (to encode positions 0 to 24)
- The Oracle has to deal with 4 possible positions (5x5), with an expectation of a single correct solution (N=25, M=1)

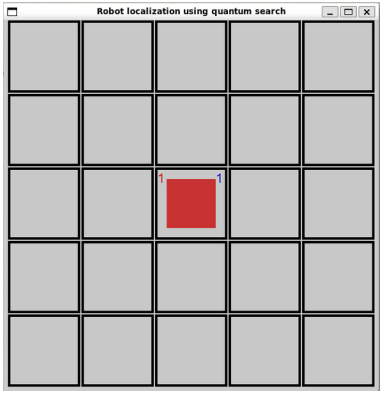

$$\text{Map qubits:} \begin{bmatrix} 0 & 0 & 0 & 0 & 0 \\ 0 & 0 & 0 & 0 & 0 \\ 0 & 0 & 1 & 0 & 0 \\ 0 & 0 & 0 & 0 & 0 \\ 0 & 0 & 0 & 0 & 0 \end{bmatrix}$$

Horizontal search qubits: [1]

Vertical search qubits: [1]

Figure 15: Simple setup with a 5x5 grid

- The suggested number of repetitions is: $R = \pi/4 \sqrt{N/M} = 4$
- The total number of required qubits is 34
- Circuit depth is 707

In order to validate results, once again, the algorithm has been tested in simulated and real quantum computers:

- using the IonQ cloud for a simulated result
- using Bluequbit for a simulated result
- in a real quantum computer from IBM.

The real IBM was the "ibm kyiv" machine, a 127qubit machine with a "2Q Error" rate (lowest two-qubit error rate) of 3.81e-3.

For the "ibm kyiv" machine, the probability to run the algorithm without any error is:

$$P \text{ (at least 1 gate fails)} = 1 - (1 - 3.81e - 3)^{707}$$
$$\approx 93\% \tag{5}$$

Once again, this value is aligned with the expected error according to [8] whose code has been integrated in the code of this paper and estimates that for this grid we would get:

- Probability of error due to noisy gates ≈ 99%
- Probability due to measurement errors = 73%

### 4.3  Grid with a pattern match with two qubits

To thoroughly assess the performance and robustness of the proposed quantum algorithm for robot localization, we designed a setup that explores the use of patterns longer than 1 qubit. In this case, a 4x5 grid will be used, with horizontal and vertical pattern searches that use two qubits. Once again, the Pygame based interface has been used to configure such a grid and to define the robot's position in a certain cell in the grid.

An obstacle has been placed in an arbitrary location, as seen in Figure 17.

The proposed double Grover search with a pattern [1, 1] in a 4x5, has to deal with a search space of 3x4 (as the length of the pattern to be searched requires two matching cells). In order to evaluate the quantum circuit, it is necessary to use:

- 4 qubits for the search space (to encode positions 0 to 12)
- The Oracle has to deal with 12 possible positions (4x3), with an expectation of a single correct solution (N=12, M=1)
- The suggested number of repetitions is: $R = \pi/4 \sqrt{N/M} = 3$

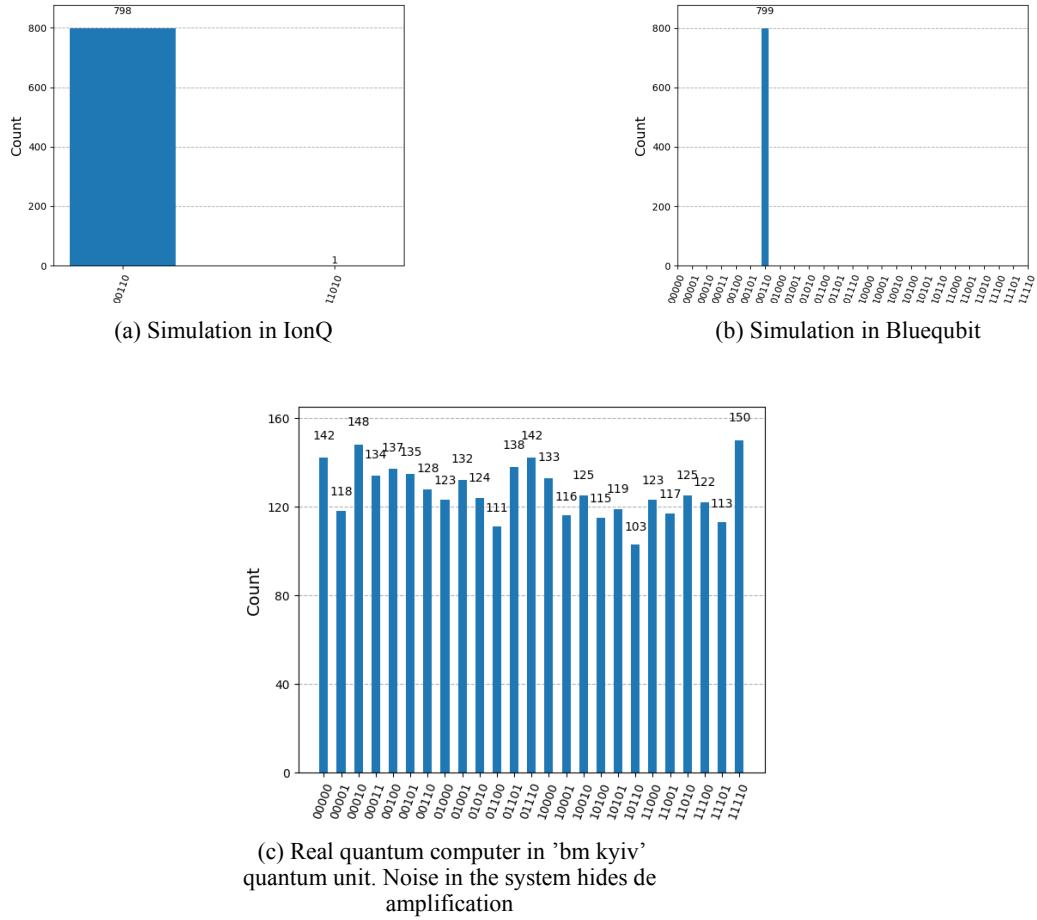

Figure 16: Histograms of the same quantum circuit obtained from simulated and real quantum computers

- The total number of required qubits is 31
- Circuit depth is 292

As this setup required 31 qubits, the system could not be tested locally on a computer. The only options available are to use the Bluequbit cloud, or run the algorithm in a real quantum computer.

The results are shown in Fig. 18.

In simulations, Grover's algorithm clearly identified the correct solution in the simulated circuits: position 10 (0101) of all possible locations, which means row 2 and column 2 of the search space (count starts from zero). But although in all the simulated tests the correct solution's probability was clearly amplified, this did not happen in the real quantum computer.

An approximation for the probability of having an erroneous result using the 'ibm kyiv' quantum computer is:

$$P \text{ (at least 1 gate fails)} = 1 - (1 - 3.81e - 3)^{292}$$
$$\approx 67\% \tag{6}$$

Using again the code from ([8]), for this $4 \times 5$ grid, we get:

- Error due to noisy gates > 99%
- Measurement errors > 50%

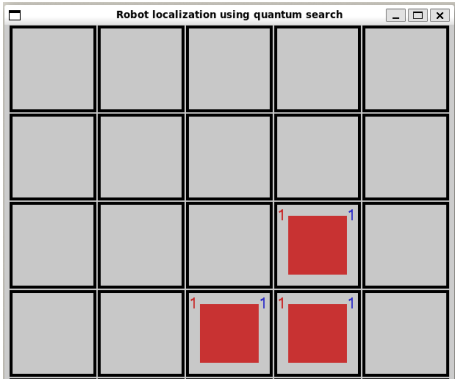

Map qubits: $\begin{bmatrix} 0 & 0 & 0 & 0 & 0 \\ 0 & 0 & 0 & 0 & 0 \\ 0 & 0 & 0 & 1 & 0 \\ 0 & 0 & 1 & 1 & 0 \end{bmatrix}$

Horizontal search qubits: [1, ¶]

Vertical search qubits: [1, ¶]

Figure 17: Setup with a 4x5 grid

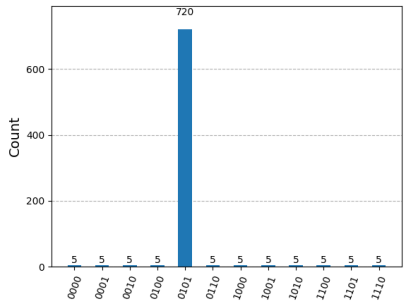

(a) Bluequbit simulation

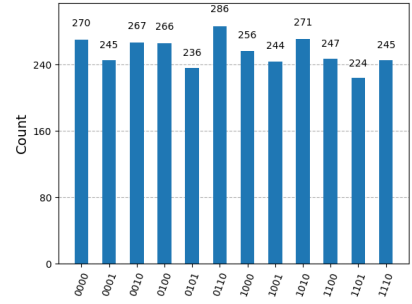

(b) Results of running the algorithm in a real 'ibm kyiv' quantum unit

Figure 18: Histograms of the same quantum circuit obtained from simulated and real quantum computers

It seems clear that current real quantum hardware lacks the robustness and scalability needed for practical applications, limiting its use in real situations.

Nevertheless, researchers are actively working to overcome these challenges, developing new quantum error correction techniques and improving qubit coherence times. Although noise mitigation techniques are a crucial aspect of reliable quantum computing, a detailed exploration of these methods is beyond the scope of this paper. The application of noise mitigation techniques, such as dynamic decoupling, noise reduction, and quantum error mitigation, requires a comprehensive understanding of quantum error correction codes, quantum control techniques, and classical algorithms.

## 5   Scalability

As mentioned, the quantum localization algorithm introduced in this paper proposes a novel method to reduce the computational complexity of searching for a robot's position in an $n \times n$ grid. As the complexity of the localization problem increases with the size of the grid (denoted as $N$), the proposed method takes advantage of an approach inspired by Grover's algorithm to reduce this complexity to $O(\sqrt{N})$.

Although the algorithm is theoretically unrestricted, its practical implementation remains limited by current engineering constraints, as demonstrated in previous experimental setups. Presently, quantum computers face capacity restrictions due to noise and the inherent difficulty of maintaining quantum states over extended periods. These disturbances can introduce errors in quantum computations, posing significant challenges in the execution of reliable operations.

However, it should be noted that this situation is expected to change in the future. Quantum gate error rates are expected to be significantly reduced in the coming years due to advances in error correction techniques, improved qubit coherence times, and enhanced fabrication processes, as shown in the trend presented in Figure 19

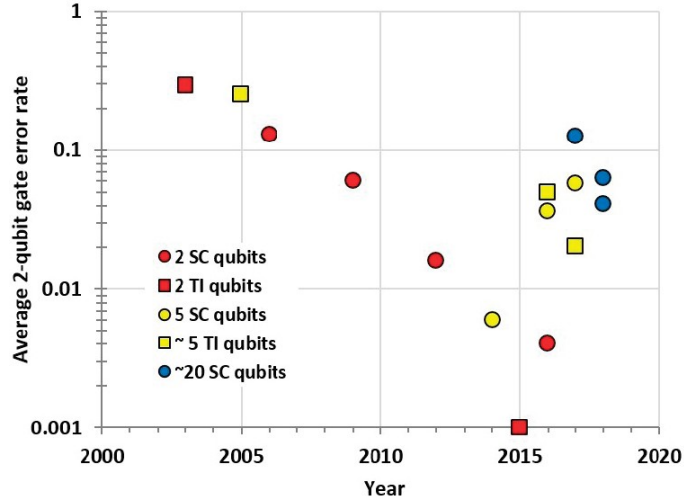

Figure 19: Recent progress in quantum gate error rates. Source: https://nap.nationalacademies.org/read/25196/chapter/9

To empirically assess the current limits of running the proposed algorithm on real quantum computers, a series of test sets have been prepared and executed in both simulated environments (locally and through third parties) and real quantum hardware provided by IBM via the IBM Quantum Platform.

These tests are presented as a simple problem of finding the position of an obstacle (marked with a "1") in a certain grid. These grids scale from 1 × 2 (2 elements) to 6 × 6 (64 elements). The results are shown in Figure 21.

As an additional comment, it should be noted that empirical testing has also shown that the complexity of the elements within the grid does not affect the performance of the search algorithm. In other words, the difficulty of finding the robot location remains the same, regardless of whether the grid contains a single distinguishing feature or multiple features. This means that the grids shown in Figure 20 require essentially the same amount of computation.

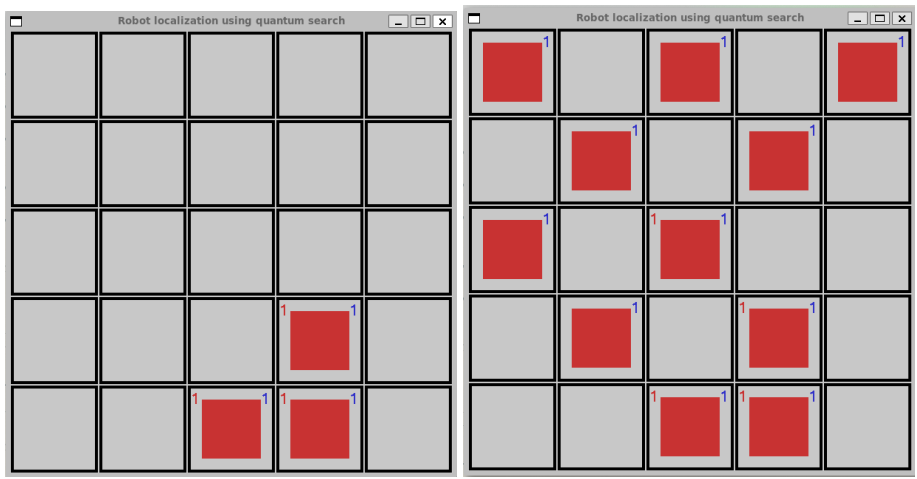

Figure 20: Grids of the same size with different amount of features require the same amount of computation

Note that for the tests shown in Figure 21, an error probability of 3.81e-3 has been considered (in line with what can be achieved in current modern quantum computers), and that the error has been calculated using the methodology proposed by [8].

## 6 Quantum advantage

Quantum advantage is the situation at which a quantum computer can solve a problem faster or more efficiently than the best classical computers available.

| SIZE | GRID | Req. qubits | Required repetitions | Circuit Depth | Approx. Error in real system | Real error : | Known Real Position: | Local Simulation (QASM) Position: | Result (Bluequbit). Position: | REAL quantum computer. Position: | Notes |
|---|---|---|---|---|---|---|---|---|---|---|---|
| 1x2 | 1 | 7 | 1 | 17 | 6% | 12% | 1 | 1 | 1 | 1 | |
| 2x2 | 00<br>01 | 10 | 2 | 59 | 20% | 12% | 3 | 3 | 3 | 3 | |
| 3x3 | 010<br>000<br>000 | 17 | 3 | 195 | 52% | 70% | 1 | 1 | 1 | 1 | (Had to repeat the job in the real computer 3 times) |
| 4x4 | 0000<br>0000<br>0100<br>0000 | 24 | 4 | 451 | 82% | 95% | 9 | 9 | 9 | -<br>(Wrong answer) | |
| 5x5 | 00000<br>00000<br>00000<br>00100<br>00000 | 34 | 4 | 707 | 93% | >99% | 18 | -<br>(Can't be calculated) | 18 | -<br>(Wrong answer) | |
| 6x6 | 000000<br>000000<br>000000<br>001000<br>000000<br>000000 | 46 | 5 | 1268 | 99% | 100% | 21 | -<br>(Can't be calculated) | -<br>(Can't be calculated) | -<br>(Wrong answer) | |

Figure 21: Scalability test results

This does not necessarily mean that the quantum computer is nowadays useful for practical applications in general: it just states that it outperforms classical systems on a specific task (in other words: although these breakthroughs mark important milestones, they do not yet necessarily translate into general-purpose, practical applications).

Quantum advantage is typically demonstrated in highly specialized problems, such as the localization problem presented in this paper.

Today in robotics, probabilistic methods are commonly employed to address the localization problem, as they can effectively combine the strengths of various localization techniques, mitigate their weaknesses, and handle uncertainty and noisy measurements, as noted in [ 10]. That is why in this paper we compare the efficiency of the proposed quantum-aided localization algorithm to a traditional probabilistic robot localization method in a grid-based scenario.

Probabilistic methods include different approaches, such as Bayesian filtering, such as Kalman filters, Extended Kalman filters, Unscented Kalman filters, Monte Carlo localization (MCL), etc. The choice of probabilistic method depends on the specific application, sensor suite, and performance requirements. Some methods, like MCL, are more computationally expensive but provide more accurate estimates, while others, like Kalman filters, are more efficient but may be less accurate.

As one of such probabilistic techniques, in this comparison Monte Carlo Localization (MCL) is used. MCL uses random sampling to represent the robot state as a set of particles. Each particle is weighted based on its likelihood, and the weights are updated over time to reflect the changing environment and sensor measurements, so for localization, the typical workflow would be:

- Initialize: Start with a uniform belief over all cells (if no prior knowledge).
- Move: If the robot is in motion, apply the motion model to predict the new state.
- Sense: Compare sensor readings with expected values and update probabilities.
- Normalization: Normalization of measured values.
- Resample: remove unlikely positions and focus on probable ones.

If we take, for example, the grid shown in Figure 17 (a 4x4 grid), we have an uncertainty area of 9 cells (since the setup is using two qubits). As MCL particles are initially spread uniformly and randomly throughout the space state, it is expected that (in our case) there are approximately 9 particles that have to: be initialized, sense their surroundings, and then resampled. As the grid grows, more particles are needed to cover the map, so for a map with N cells, the complexity of localization increases with $O(N)$.

Based on the initial code proposal of [38], we simulate the configuration (described in Figure 17). Figure 22 shows the evolution of the MCL algorithm to estimate robot location if starting from an unknown location.

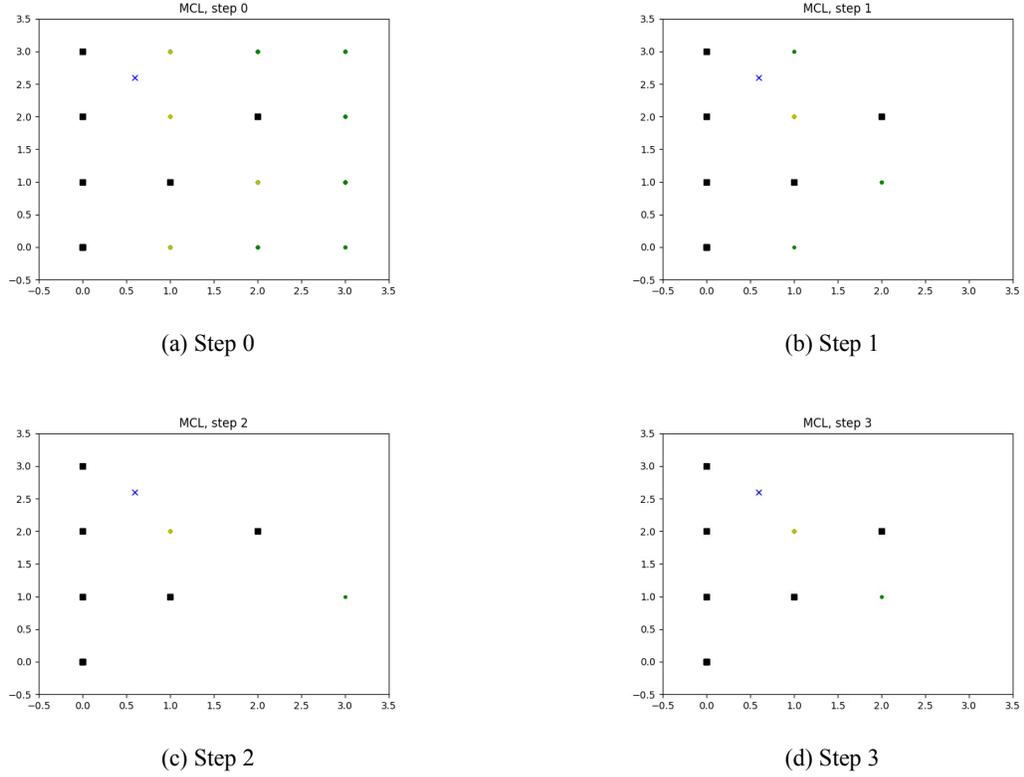

(a) Step 0          (b) Step 1

(c) Step 2          (d) Step 3

Figure 22: Iterations of MCL in the setup described in 17. Black squares represent obstacles, green dots the particles (in light green the resampled particles), and the blue circle represents the robot

Figure 22 shows how the particle filter iteratively approaches to the solution, starting from a set of randomly distributed particles.

In our proposed approach, since Grøver's algorithm requires repeating certain steps according to the size of the state map ($N$) following $Repeat = \pi/4 \sqrt{N}$, complexity increases with the square root of N.

This means that for large values of N, O(sqrt(N)) grows much slower than O(N). This means that the second method will be significantly faster for large inputs.

To illustrate this, consider an example:

- If the size of the map is $N = 100$, then $\sqrt{N} \approx 10$ for the quantum-aided approach, and $N = 100$ for classical methods.
- If $N = 1{,}000{,}000$, then $\sqrt{N} \approx 1{,}000$, and $N = 1{,}000{,}000$.

As N grows, the relative difference between O(N) and O(sqrt(N)) becomes more pronounced, and O(N) grows much faster. This implies that for larger maps the proposed quantum-aided approach requires much fewer calculations (obviously, assuming there exists a real quantum device that can compute such calculations).

# 7  Conclusions and Future Work

The experimental results obtained from the proposed quantum algorithm for robot localization demonstrate its feasibility and effectiveness in solving the problem. The approach successfully localizes the robot on a 2D map, confirming the theoretical validity of the quantum-based method. However, the performance of the algorithm on real quantum computers is hindered by current technological limitations. As the size of the map increases, the accuracy of the algorithm degrades as a result of the difficulties in maintaining the fragile quantum state for extended periods. This is a common challenge faced by many quantum algorithms, and our results highlight the need for advances in quantum hardware to fully exploit the potential of quantum computing.

Furthermore, our simulations on classical computers reveal another limitation: the number of qubits required to simulate the algorithm grows exponentially with the size of the map. Currently, we are unable to simulate the algorithm for maps that require more than 40 qubits, which restricts the scalability of our approach. The limitations of current quantum systems underscore the need for developing more efficient quantum algorithms and simulation methods. These advancements will help bridge the gap until true quantum computers can effectively utilize a larger number of qubits.

Fortunately, recent advances in the design and manufacturing of quantum computers indicate a promising trajectory, with many of these current shortcomings expected to be addressed within the next decade.

Additionally, coherence time (the amount of time that a qubit can maintain its quantum state, or coherence, without being disrupted or affected by its environment) is also being increased following a similar trend. This means that longer (or more complex) calculations can be executed without errors in a certain algorithm.

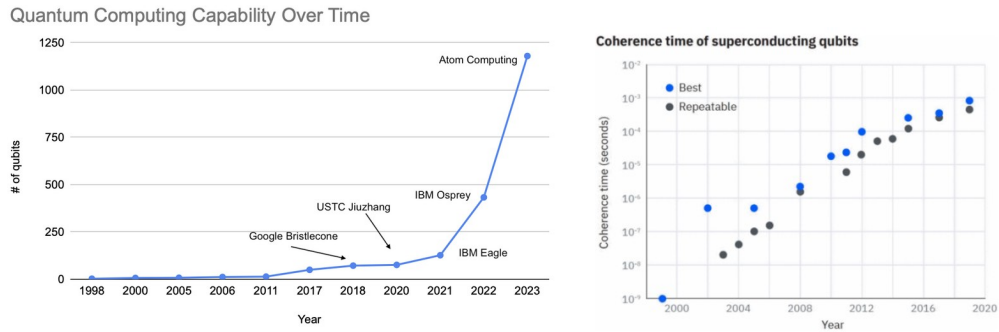

Figure 23: Available qubits per chip in some of the main quantum computer providers (left) and evolution of coherence times (right). Sources: https://www.netmeister.org/blog/pqc-2024-01.html and Eindhoven University

However, despite the promising results, there are several research lines for future work to improve the performance and applicability of the quantum algorithm proposed in this paper. Work on this paper has been focused in the design and validation of the algorithm, but other aspects can be developed in the future, such as:

- Non-exact match quantum algorithm: One potential direction is to explore non-exact match quantum algorithms, such as those based on the Hamming distance search. This could allow for more robust and flexible localization, especially in the presence of noisy or incomplete data.
- Alternative algorithms beyond Grover's algorithm: Our current implementation relies on Grover's algorithm, which may not be the most efficient or suitable choice for this problem. Investigating other quantum algorithms, such as the Quantum Approximate Optimization Algorithm (QAOA) or the Variational Quantum Eigensolver (VQE), could lead to improved performance and reduced resource requirements.
- Quantum error correction and noise mitigation: To overcome the challenges posed by current quantum hardware, research and implementation of quantum error correction techniques or noise mitigation strategies could help maintain the coherence of the quantum state for longer periods.
- Hybrid quantum-classical approaches: Exploring hybrid approaches that combine the strengths of quantum and classical computing could provide a more practical and efficient solution for robot localization. This might involve using quantum computing for specific subtasks or exploiting classical machine learning techniques to supplement the quantum algorithm.

By addressing these research directions, we can further improve the performance and applicability of the proposed quantum algorithm, ultimately paving the way for future practical and efficient solutions to the robot localization problem.

During the preparation of this work, the author used Llama3.1 generative AI in order to revise the spelling, grammar errors and for rephrasing sentences for clarity. After using this tool/service, the author reviewed and edited the content as needed and assumes full responsibility for the content of the publication.

This research was funded by the ELKARTEK Research Program of the Basque Government, project #KK-2024/00024. The APC was funded by Fundación TECNALIA Research and Innovation.

The authors declare no conflict of interest. The funders had no role in the design of the study; in the collection, analyses, or interpretation of data; in the writing of the manuscript; or in the decision to publish the results.